\documentclass[conference]{IEEEtran}
\IEEEoverridecommandlockouts
\usepackage{cite}
\usepackage{amsmath,amssymb,amsfonts}
\usepackage{graphicx}
\usepackage{textcomp}

\usepackage[table]{xcolor}% http://ctan.org/pkg/xcolor
\usepackage{times}
\usepackage{soul}
\usepackage{url}
\usepackage[hidelinks]{hyperref}
\usepackage[utf8]{inputenc}
\usepackage[small]{caption}
\usepackage{tablefootnote}
\usepackage{graphicx}
\usepackage{amsmath}
\usepackage{amsfonts}
\usepackage{amsthm}
\usepackage{booktabs}
\usepackage{algorithm}
\usepackage{algorithmicx}
\usepackage{algpseudocode}
\usepackage[normalem]{ulem}

\usepackage{subcaption}
\useunder{\uline}{\ul}{}
\def\BibTeX{{\rm B\kern-.05em{\sc i\kern-.025em b}\kern-.08em
    T\kern-.1667em\lower.7ex\hbox{E}\kern-.125emX}}
\begin{document}

\title{M-CELS: Counterfactual Explanation for Multivariate Time Series Data Guided by Learned Saliency Maps}

% \author{
%     \IEEEauthorblockN{ 
%     Peiyu Li, 
%     Souka\"ina Filali Boubrahimi,
%     Shah Muhammad Hamdi
%     }
    
%     \IEEEauthorblockA{Department of Computer Science,
%  Utah State University,
%  Logan, UT 84322, USA
%     \\ } Emails: \texttt{\{peiyu.li, soukaina.boubrahimi, s.hamdi\}@usu.edu} }

% \author{\IEEEauthorblockN{Peiyu Li}
% \IEEEauthorblockA{\textit{Department of Computer Science} \\
% \textit{Utah State University}\\
%  Logan, UT 84322, USA \\
% peiyu.li@usu.edu}
% \and
% \IEEEauthorblockN{Souka\"ina Filali Boubrahimi,}
% \IEEEauthorblockA{\textit{Department of Computer Science} \\
% \textit{Utah State University}\\
%  Logan, UT 84322, USA \\
% soukaina.boubrahimi@usu.edu}
% \and
% \IEEEauthorblockN{Shah Muhammad Hamdi}
% \IEEEauthorblockA{\textit{Department of Computer Science} \\
% \textit{Utah State University}\\
%  Logan, UT 84322, USA \\
% s.hamdi@usu.edu}
% }

\author{
    \IEEEauthorblockN{ 
    Peiyu Li,
    Omar Bahri,  
    Souka\"ina Filali Boubrahimi,
    Shah Muhammad Hamdi
    }
    
    \IEEEauthorblockA{Department of Computer Science,
 Utah State University,
 Logan, UT 84322, USA
    \\ } Emails: \texttt{\{peiyu.li, omar.bahri, soukaina.boubrahimi, s.hamdi\}@usu.edu} }

% \and
% \IEEEauthorblockN{2\textsuperscript{nd} Anonymous}
% \and
% \IEEEauthorblockN{3\textsuperscript{rd} Anonymous}
% \and
% \IEEEauthorblockN{4\textsuperscript{th} Anonymous}
% \and
% \IEEEauthorblockN{5\textsuperscript{th} Anonymous}
% \and
% \IEEEauthorblockN{6\textsuperscript{th} Anonymous}

\maketitle

\begin{abstract}
Over the past decade, multivariate time series classification has received great attention. Machine learning (ML) models for multivariate time series classification have made significant strides and achieved impressive success in a wide range of applications and tasks. The challenge of many state-of-the-art ML models is a lack of transparency and interpretability. In this work, we introduce M-CELS, a counterfactual explanation model designed to enhance interpretability in multidimensional time series classification tasks. Our experimental validation involves comparing M-CELS with leading state-of-the-art baselines, utilizing seven real-world time-series datasets from the UEA repository. The results demonstrate the superior performance of M-CELS in terms of validity, proximity, and sparsity, reinforcing its effectiveness in providing transparent insights into the decisions of machine learning models applied to multivariate time series data. 
\end{abstract}


\section{Introduction}
\label{sec: intro}
In recent years, there has been widespread adoption of time series classification methods across various domains \cite{eskandarinasab2024enhancing}, \cite{hosseinzadeh2023metforc}. The success of data-driven time series classification methods is largely attributed to the availability of large-scale data, which has facilitated the development of highly accurate models for real-world applications. However, most of the time series model development efforts, especially deep learning models \cite{ismail2019deep}, consider accuracy as their foremost priority without necessarily providing a tracing mechanism of the model's decision-making process which limits their interpretability. This lack of interpretability poses a challenge, as scientific communities often require explanations to support their trust in the models. Therefore, the establishment of trust between humans and decision-making systems has become critically important. This can be achieved by incorporating interpretable explanations during the model development phase or by developing post-hoc explanation solutions. By doing so, we can enhance transparency and foster trust in decision-making processes, particularly in critical domains where human judgment and accountability are paramount.
% Domain experts are hesitant to rely solely on machine-driven decisions and instead seek insights into why a particular treatment or course of action is recommended \cite{adadi2018peeking}. For example, in life-changing decisions such as critical disease management, it is important to know the reasons behind choosing a treatment line over another one \cite{kundu2021ai}. The consequences of a poor decision can be significant, not only in terms of potential loss of life but also substantial financial implications \cite{xu2019deep}. 
To address concerns regarding the lack of transparency in machine learning models, experts from different domains are proposing EXplainable Artificial Intelligence (XAI) methods to provide trustworthy explanations of the decision-making processes of machine learning models \cite{tjoa2020survey}. XAI methods aim to provide reliable explanations for the decision-making processes of these models while maintaining high levels of performance. There are two dominant paradigms of XAI: intrinsic interpretable and post-hoc explanation methods for opaque models \cite{zhou2021s}. Examples of intrinsic interpretable models include logistic regression and decision tree-based induction, which can provide an intrinsic explanation by their simple structure models. On the other hand, post-hoc explanation methods can be further categorized into feature attribution methods and counterfactual explanation methods. Feature attribution methods aim to explain the rationale behind a model's decision, while counterfactual explanation methods identify the smallest input modifications that would result in a different decision. In this paper, we focus only on post-hoc XAI methods, which have gained significant attention in the field. These methods play a crucial role in enhancing the interpretability of opaque models, enabling stakeholders to gain a better understanding of the factors driving the models' predictions.

In recent years, significant research advancements have been accomplished on explainability in the computer vision and natural language processing (NLP) domains, but there are still many challenges to be addressed to provide interpretability for the time series domain \cite{nguyen2020model}.  A lot of efforts have been made to provide post-hoc XAI for image \cite{vermeire2022explainable}, vector-represented data \cite{pawelczyk2020learning} and univariate time series data \cite{delaney2021instance}, \cite{li2022motif}, while significantly less attention has been paid to multivariate time series data \cite{ates2021counterfactual}. 

In this work, we aim to fill this gap by introducing M-CELS, a novel extension of counterfactual explanations via learned saliency maps on univariate time series to enhance transparency in multivariate time series classification models. To the best of our knowledge, this is the first effort to learn a saliency map specifically for the purpose of producing high-quality counterfactual explanations for multivariate time series data. 

% The rest of this paper is organized as follows: In section II, we lay the ground for our research by introducing the related works. Section III describes optimal properties that a good counterfactual explanation should satisfy. In Section IV, we introduce the background of CELS, and in Section V we introduce M-CELS in detail. We present the experimental results and evaluations in comparison to other baselines in Section VI. In section VII, we conduct the ablation study. Finally, we conclude our work in Section VIII.

\section{Related work}
In the post-hoc interpretability paradigm, the wCF method by Wachter et al. \cite{wachter2017counterfactual} stands as a prominent approach that first introduced the counterfactual explanation. wCF generates counterfactuals by minimizing a loss function to alter decision outcomes while preserving the minimum Manhattan distance from the original input instance. Extensions like TimeX \cite{filali2022mining} and SG-CF \cite{li2022sg} have augmented counterfactual quality by integrating extra terms into the loss function. TimeX introduces a Dynamic Barycenter Average Loss term for a more contiguous counterfactual, while SG-CF leverages previously mined shapelets for better interpretability. Nonetheless, these methods are primarily designed for univariate time series data, posing challenges for direct application to multivariate time series.

NG-CF employs Dynamic Barycenter (DBA) averaging of the query time series and its nearest unlike neighbor for counterfactual generation. Recent methods like those employing mined shapelets, temporal rules or Matrix Profile \cite{li2022motif, bahri2022shapelet, bahri2022temporal, bahri2024discord} aim to enhance interpretability. However, they often encounter limitations of extensive processing time for shapelet or rule extraction.

CoMTE \cite{ates2021counterfactual}, proposed by Ates et al., targets generating counterfactuals for multivariate time series by evaluating the impact of turning off one variable at a time. However, CoMTE tends to generate low sparsity as it modifies the entire time series data \cite{bahri2022temporal}. Another approach, AB-CF proposed by \cite{li2023attention}, underscores minimum discriminative contiguous segment replacement for sparser and higher-validity counterfactuals. Nonetheless, it presupposes the presence of discriminative segments across all dimensions within the same time-step interval.

In addressing these challenges, our work introduces optimization techniques tailored to multivariate time series, guiding perturbations through saliency maps. Our contributions are summarized below:
\begin{enumerate}
   \item \textbf{Optimization Techniques for Multivariate Time Series}: We introduce novel optimization methods specifically tailored to the complexities of multivariate time series data. These techniques guide the generation of counterfactuals, ensuring they are both meaningful and interpretable across multiple dimensions.

   \item \textbf{Use of Saliency Maps for Enhanced Sparsity}: By leveraging saliency maps, our approach identifies and targets the most influential features within the time series data. This focused perturbation promotes higher sparsity. As a result, fewer changes are made to the original time series, making the counterfactuals more actionable for users.

   \item \textbf{Improved Interpretability in Time Series Classification}: The combination of targeted perturbations and enhanced sparsity directly contributes to better interpretability in time series classification tasks. Users can more easily discern the rationale behind model decisions and the influence of different variables over time.
\end{enumerate}

\section{Background Works: CELS}
Recently, the CELS model \cite{li2023cels} introduced the concept of generating counterfactual explanations for univariate time series classification using learned saliency maps. Building on this approach, we present M-CELS, an extension of CELS, to generate high-sparsity and high-validity counterfactual explanations for multivariate time series data. Our methodology adapts the principles of CELS to handle the added complexity of multivariate datasets, ensuring meaningful and interpretable counterfactuals across multiple dimensions. In this section, we briefly describe the CELS background and the main components we utilized in M-CELS.

\subsection{Problem definition for CELS}
CELS \cite{li2023cels} is a novel counterfactual explanation method originally designed for univariate time series data, offering high proximity, sparsity, and interpretability in post-hoc explanations for time series classification.

Given a set of $N$ univariate time series $\mathcal{D}=$ $\left\{\mathbf{x}_1, \ldots, \mathbf{x}_N\right\}$ and a univariate time series classification model $f: \mathbb{X} \rightarrow \mathcal{Y}$, where $\mathbb{X} \in \mathbb{R}^T$ is the $T$-dimensional feature space and $\mathcal{Y}=\{1,2, \ldots, C\}$ is the label space with $C$ classes. For an instance-of-interest $\mathbf{x} \in \mathcal{D}$, a time series with $T$ time steps along with a predicted probability distribution $\hat{\mathbf{y}}=f(\mathbf{x})$ over $C$ classes where $\hat{\mathbf{y}} \in[0,1]^C$ and $\sum_{i=1}^C \hat{\mathbf{y}}_i=1$. The top predicted class $z$ where $z=\operatorname{argmax} \hat{\mathbf{y}}$ is the class of interest for which a counterfactual explanation is sought. The instance of interest $\mathbf{x}$ is defined as the original query instance, and the perturbed result $\mathbf{x}'$ is defined as the counterfactual instance or counterfactual.

The goal of CELS is to learn a saliency map $\boldsymbol{\theta} \in [0,1]^T$ for class $z$, where each element represents the importance of its corresponding time step. Values close to 1 indicate strong evidence for class $z$, while values near 0 indicate no importance. CELS assumes that the importance of a time step reflects how much $P(\hat{\mathbf{y}}_z \mid \mathbf{x})$ changes when $\mathbf{x}$ is perturbed to $\mathbf{x}'$. A successfully learned saliency map $\boldsymbol{\theta}$ will assign high values to the regions in a time series where the perturbation function would shift the model's predictions away from $z$. 

CELS contains three key components that work together: (1) the Nearest Unlike Neighbor Replacement strategy learns which time series from the background dataset $\mathcal{D}$ are the best for replacement-based perturbation. (2) the Learning Explanation aims to learn a saliency map $\boldsymbol{\theta}$ for the instance of interest by highlighting the most important time steps that provide evidence for a model prediction. (3) the Saliency Map Guided Counterfactual Perturbation function which perturbs the instance of interest $\mathbf{x}$ to force the perturbation shift the model's prediction away from the class of interest $z$.
\subsection{Replacement Strategy: Nearest Unlike Neighbor (nun)}
NG \cite{delaney2021instance} generates in-distribution perturbations by replacing time steps with those from the nearest unlike neighbor in a background dataset $\mathcal{D}$. Inspired by NG, CELS similarly selects the nearest unlike neighbor time series from $\mathcal{D}$ (usually the training dataset) to guide the perturbation of the instance $\mathbf{x}$. The nearest unlike neighbor is defined as the instance $nun$ from a different class $z'$, with the smallest Euclidean distance to $\mathbf{x}$. In binary classification, $z'$ is the opposite class of $z$, while in multi-class classification, it is the class with the second-highest probability. This approach helps explain model predictions and generates counterfactual explanations by identifying the minimal changes needed to alter the prediction \cite{delaney2021instance}.

\subsection{Learning Explanation}
Saliency values $\mathbb{\theta} \in \mathbb{R}^{1 \times T}$ are learned using a novel loss function (Equation \ref{eq: total loss}) including three key components. 

First, a $L_{\operatorname{Max}}$ is designed to promote the perturbation function to generate high-validity counterfactual instances. $P\left(\hat{\mathbf{y}}_{z'} \mid \mathbf{x}'\right)$ represents the class probability of class $z'$ (the target class) predicted by the classification model $f$ for the perturbated counterfactual instance $\mathbf{x}'$. Specifically, $L_{\operatorname{Max}}$ is incorporated to maximize the class probability of class $z'$ predicted by the classification model $f$ for the perturbated counterfactual instance $\mathbf{x}'$, if $P\left(\hat{\mathbf{y}}_{z'} \mid \mathbf{x}'\right)$ is optimized to be close to 1, then we can get high validity counterfactual explanation:

\begin{equation}
    L_{\operatorname{Max}}= 1-P\left(\hat{\mathbf{y}}_{z'} \mid \mathbf{x}'\right)
\end{equation}

Next, to encourage simple explanations with minimal salient time steps, the $L_{\text {Budget }}$ loss is introduced. This component promotes the values of $\boldsymbol{\theta}$ to be as small as possible, where intuitively, values that are not important should be close to 0 according to the defined problem:

\begin{equation}
    L_{\text {Budget }}=\frac{1}{T} \sum_{t=1}^T\boldsymbol{\theta}_{ t}
\end{equation}

The saliency map is then encouraged to be temporally coherent, where neighboring time steps should generally have similar importance. To achieve this coherence, a time series regularizer $L_{\text {TReg }}$ is introduced, minimizing the squared difference between neighboring saliency values:

\begin{equation}
    L_{\mathrm{TReg}}=\frac{1}{T} \sum_{t=1}^{T-1}\left(\boldsymbol{\theta}_{t}-\boldsymbol{\theta}_{t+1}\right)^2
\end{equation}

Finally, all the loss terms are summed, with $L_{\operatorname{Max}}$ being scaled by a $\lambda$ coefficient to balance the three components in counterfactual explanation behavior. 
\begin{equation}
\label{eq: total loss}
    L(P(\hat{\mathbf{y}} \mid \mathbf{x}) ; \boldsymbol{\theta})=\lambda *L_{\mathrm{Max}}+L_{\mathrm{Budget}}+L_{\mathrm{TReg}}
\end{equation}

\subsection{Counterfactual Perturbation}
A saliency map is derived for the top predicted class $z$ to identify the key time steps contributing to the prediction, which also plays an important role in guiding the generation of counterfactual explanations. CELS perturbs the instance $\mathbf{x}$ by replacing its important time steps with those from the nearest unlike neighbor $nun$, as guided by the saliency map $\boldsymbol{\theta}$. The counterfactual perturbation function is defined as:

\begin{equation}
\label{eq:perturb}
    \mathbf{x}'=  \mathbf{x} \odot \left(1-\boldsymbol{\theta}\right) +  nun \odot \boldsymbol{\theta}
\end{equation}
Specifically, if $\boldsymbol{\theta}_t = 0$, which means that the time step $t$ is not important for the prediction decision on class $z$, then the original time step $\mathbf{x}_t$ remains unchanged. In other words, if $\boldsymbol{\theta}_t = 1$, which means that the time steps $t$ provide important evidence for the prediction decision on class $z$, then $\mathbf{x}_t$ is replaced with the corresponding time step of nearest unlike neighbor $nun$. In summary, the perturbation function generates the counterfactual explanation $\mathbf{x}'$ by performing time step interpolation between the original time steps of $\mathbf{x}$ and the replacement series $nun$.

\section{M-CELS}
The Nearest Unlike Neighbor replacement strategy and the counterfactual perturbation function, initially designed for univariate time series data, seamlessly extend to multivariate time series data with direct applicability. The key adaptation lies in the saliency map learning process. When addressing multi-dimensional time series data, the conventional approach of learning a one-dimensional saliency map is replaced by the exploration of a multi-dimensional saliency map, denoted as $\boldsymbol{\theta} \in$ $[0,1]^{T\times D}$, capturing the nuances of the multi-dimensional temporal sequence. This section introduces the notation and problem definition for counterfactual explanation using learned saliency maps on multivariate time series data. It elucidates the saliency map learning process's significant deviation from its univariate counterpart, providing a comprehensive understanding of the intricate adjustments required for effective application in the multivariate context.
\subsection{Notation}
\textit{We assume a univariate time series $\textbf{x} = \{{x}_1, {x}_2, ..., {x}_\textit{T}\}$ is an ordered set of real values, where \textit{T} is the length of the time series. In the case of multivariate time series, the time series is a list of vectors over $\textit{D}$ dimensions and $\textit{T}$ observations, $\mathbf{x} = [\textbf{x}^1, \textbf{x}^2, ..., \textbf{x}^\textit{D}]$. Then we can define a multivariate time series dataset $\mathcal{D}=\{\mathbf{x}_{0},\mathbf{x}_{1},..., \mathbf{x}_{n}\}$ as a collection of n multivariate time series. Assume we are given a multivariate time series classification model $f:\mathbf{x} \rightarrow \mathcal{Y}$, where $\mathbf{x}\in \mathbb{R}^{\textit{T}\times \textit{D}}$ is the feature space of the input multivariate time series, where each multivariate time series has mapped to a mutually exclusive set of classes $\mathcal{Y}=\{1,2, \ldots, C\}$ }, where $C$ is the number of classes in dataset $\mathcal{D}$.

\subsection{Problem Definition. }
Let us consider an instance-of-interest $\mathbf{x} \in \mathcal{D}$, a multivariate time series with $D$ dimensions and $T$ time steps per dimension along with a predicted probability distribution $\hat{\mathbf{y}}=f(\mathbf{x})$ over $C$ classes where $\hat{\mathbf{y}} \in[0,1]^C$ and $\sum_{i=1}^C \hat{\mathbf{y}}_i=1$. The top predicted class $z$ where $z=\operatorname{argmax} \hat{\mathbf{y}}$ has predicted confidence $\hat{\mathbf{y}}_{\mathbf{z}}$ is the class of interest for which we seek a counterfactual explanation. 

Our goal is to learn a saliency map $\boldsymbol{\theta} \in$ $[0,1]^{D\times T}$ for the class of interest $z$ where each element represents the importance of a corresponding time step at a corresponding dimension at the very first step. Values in $\boldsymbol{\theta}$ close to 1 indicate strong evidence for class $z$, while values in $\boldsymbol{\theta}$ close to 0 indicate no importance. We assume the importance of a time step $t$ should reflect the expected scale of the change of $P\left(\hat{\mathbf{y}}_z \mid \mathbf{x}\right)$ when $\mathbf{x}$ is perturbed: $\left|P\left(\hat{\mathbf{y}}_z \mid \mathbf{x}\right)-P\left(\hat{\mathbf{y}}_z \mid \mathbf{x}' \right)\right|$, where $\mathbf{x}'$ is perturbed version of $\mathbf{x}$. Then we design a perturbation function based on the learned saliency map to generate a counterfactual explanation $\mathbf{x}'$ for the class of interest $z$. A successfully learned saliency map $\boldsymbol{\theta}$ will assign high values to the regions in a time series where the perturbation function would shift the model's predictions away from $z$. We define the instance of interest $\mathbf{x}$ for which we seek a counterfactual explanation as the original query instance and the perturbed result $\mathbf{x}'$ as the counterfactual instance or counterfactual.

\subsection{Learning Explanation for Multivariate Time Series Data}
In the case of multivariate time series data, the learning explanation would be different from the process on univariate time series data. Saliency values $\mathbb{\theta} \in \mathbb{R}^{T \times D}$ are learned where $\theta_{t, d}$ is the saliency value of dimension d at time step t. The loss function utilized to optimize saliency maps for deriving an intuitive instance-based counterfactual explanation, denoted as $\mathbf{x}$, undergoes distinct adaptations.

The $L_{\operatorname{MMax}}$ remains unchanged to ensure the high validity of the counterfactual explanation generation, where $z'$ is the target class of the counterfactual explanation.
\begin{equation}
    L_{\operatorname{MMax}}=\left(1-P\left(\hat{\mathbf{y}}_{z'} \mid \mathbf{x}'\right)\right)
\end{equation}
However, approaches to calculating the Budget loss and TV norm loss differ. Equation \ref{eq: mbudget} introduces the multivariate budget loss, computing the average absolute saliency values across dimensions and time steps. Enforcing sparsity in the saliency map offers insights into the overall perturbation magnitude, ensuring judicious distribution of changes across dimensions.
\begin{equation}\label{eq: mbudget}
    L_{\mathrm{MBudget}}=\frac{1}{D} \sum_{d=1}^{D}\left(\frac{1}{T} \sum_{t=1}^T\boldsymbol{\theta}_{t, d}\right)
\end{equation}
Equation \ref{eq: mtreg} presents the multivariate temporal regularization term, capturing the smoothness and temporal consistency of saliency maps across dimensions. It discourages abrupt changes in saliency values over time, promoting coherent explanations across consecutive time steps.
\begin{equation}\label{eq: mtreg}
    L_{\mathrm{MTReg}}=\frac{1}{D} \sum_{d=1}^D\left(\frac{1}{T} \sum_{t=1}^{T-1}\left(\theta_{t, d}-\theta_{t+1, d}\right)^2\right)
\end{equation}
Finally, the overall loss function in Equation \ref{eq: newloss} integrates the three components ($L_{\mathrm{MMax}}$, $L_{\mathrm{MBudget}}$, $L_{\mathrm{MTReg}}$), reflecting diverse objectives in the multivariate time series setting. These equations collectively establish a comprehensive framework tailored for the nuanced characteristics of multivariate time series data, ensuring an effective and meaningful learning explanation process.
\begin{equation} \label{eq: newloss}
    L(P(\hat{\mathbf{y}} \mid \mathbf{x}) ; \boldsymbol{\theta})=\lambda *L_{\mathrm{MMax}}+L_{\mathrm{MBudget}}+L_{\mathrm{MTReg}}
\end{equation}
     
% \begin{figure*}[ht!]
% \centering
% \includegraphics[scale = .5]{Images/MCELS.pdf}
% \caption{The architecture of M-CELS (Counterfactual Explanation for Multivariate Time Series Data Guided by Learned Saliency Map)}
% \label{Fig:mainfig}
% \end{figure*}

% \section{Saliency map normalization}
% In Algorithm \ref{alg:alg1} line 12, there is a saliency map normalization after the saliency map learning. This normalization was initially designed to generate more sparse counterfactuals by ignoring the low saliency score. In CELS, the normalization is to set the saliency score to 1 if the saliency score is larger than a threshold $k$ and 0 if the saliency score is larger than a threshold. We 
% \begin{figure}[htpb!]
% \centering
% \includegraphics[scale = .55]{Images/parameter study/lambda1 (1).pdf}
% \caption{Parameter study: explore the threshold }
% \label{Fig:mainfig}
% \end{figure}

\begin{algorithm}
\caption{M-CELS Algorithm\label{alg:alg1}} 
\textbf{Inputs:} Instance of interest $\mathbf{x}$, pretrained time series classification model $f$, background dataset $\mathcal{D}$, the total number of time steps $T$ of the instance for each dimension, the number of dimensions $D$, a threshold $k$ to normalize $\boldsymbol{\theta}$ at the end of learning.\\
\textbf{Output:} Saliency map $\boldsymbol{\theta}$ for instance of interest $\mathbf{x}$, counterfactual explanation $\mathbf{x}'.$
\begin{algorithmic}[1]
  \State $\boldsymbol{\theta}$ $\leftarrow$ random\_uniform(low = 0,high = 1)
  \State $nun, z'$ $\leftarrow$ Nearest Unlike Neighbor($\mathbf{x}$, $\mathcal{D}$, $f$) 
  \For{epoch $\leftarrow$ 1 to epochs}
      \State $\mathbf{x}'= \mathbf{x} \odot \left(1-\boldsymbol{\theta}\right) +  nun \odot \boldsymbol{\theta}$
      \State $ L_{\operatorname{MMax}}=1-P\left(\hat{\mathbf{y}}_{z'} \mid \mathbf{x}'\right)$
      \State $L_{\mathrm{MBudget}}=\frac{1}{D} \sum_{d=1}^{D}\left(\frac{1}{T} \sum_{t=1}^T\boldsymbol{\theta}_{t, d}\right)$
      \Comment{$\theta_{t, d}$ is the saliency value of dimension d at time step t}
      \State $L_{\mathrm{MTReg}}=\frac{1}{D} \sum_{d=1}^D\left(\frac{1}{T} \sum_{t=1}^{T-1}\left(\theta_{t, d}-\theta_{t+1, d}\right)^2\right)$
      \State $L(P(\hat{\mathbf{y}} \mid \mathbf{x}) ; \boldsymbol{\theta})=\lambda * L_{\mathrm{MMax}}+L_{\mathrm{MBudget}}+L_{\mathrm{MTReg}}$
      \State $\boldsymbol{\theta} \leftarrow \boldsymbol{\theta} - \eta \nabla J(\boldsymbol{\theta})$
      \State clamp($\boldsymbol{\theta}$, low = 0, high = 1)   
  \EndFor
\State $\boldsymbol{\theta}_{t, d} = 
\begin{cases} 
\boldsymbol{\theta}_{t, d} & \text{if } \boldsymbol{\theta}_{t, d} > k \\
0 & \text{otherwise}
\end{cases}$
\Comment{normalize the saliency map to get the final counterfactual explanation}
\State $\mathbf{x}'= \mathbf{x} \odot \left(1-\boldsymbol{\theta}\right) +  nun \odot \boldsymbol{\theta}$
\State \Return {$\boldsymbol{\theta}$, $\mathbf{x}'$}
  
\end{algorithmic}
\end{algorithm}
\section{Experimental Study}
% \begin{figure*}[t!]
% \centering
% \includegraphics[scale = .35]{images/alg_fig.png}
% \caption{Counterfactual explanation for time series data via learned saliency maps}
% \label{Fig:mainfig}
% \end{figure*}
\subsection{Baseline Methods}
We evaluated our proposed M-CELS model with the other three baselines, Alibi, Native guide counterfactual (NG), and Attention-based counterfactual (AB).
\begin{itemize}
    \item \textbf{Alibi Counterfactual (Alibi):} Alibi follows the work
of Wachter et al. \cite{wachter2017counterfactual}, which constructs counterfactual explanations by optimizing an objective function, 
    \begin{equation}
        L = L_{pred} + L_{L1},
    \end{equation}
     where $L_{pred}$ guides the search towards points $\textbf{x}^{\prime}$ which would change the model prediction and $L_{L1}$ ensures that $\textbf{x}^{\prime}$ is close to $\textbf{x}$. 
    
    \item \textbf{ Native guide counterfactual (NG-CF):} NG-CF uses Dynamic Barycenter (DBA) averaging of the query time series $\textbf{X}$ and the nearest unlike neighbor from another class to generate the counterfactual example \cite{delaney2021instance}.

    \item \textbf{Attention-based Counterfactual Explanation (AB-CF):} AB-CF leverages the Shannon entropy to extract the most important $k$ subsequences by measuring the information embedded in each segment given the probability distribution and introducing the nearest unlike neighbor replacement to generate the counterfactual example \cite{li2023attention}.

\end{itemize}

\subsection{Dataset}
Building upon the investigation conducted by Li et al. \cite{li2023attention}, we conducted an assessment of our proposed methodology using identical benchmarks. The evaluation was performed on the seven publicly available multivariate time series datasets sourced from the University of East Anglia (UEA) MTS archive \cite{bagnall2018uea}. Table \ref{tab:UEA} presents the metadata details for these seven datasets. 
\begin{table}[!t]
\centering
\caption{UEA datasets metadata}
\begin{tabular}{|c||c|c|c|c|c|c|c}
\hline
\label{tab:UEA}
ID & Dataset Name   & TS length &  Dimensions & classes\\ \hline
0  & ArticularyWordRecognition         & 144     &9 &25\\ \hline
1  & BasicMotions         & 100       &6 & 4\\ \hline
2  & Cricket       & 1197        &6 & 12\\ \hline
3  & Epilepsy      & 206        &3 &4\\ \hline
4  & ERing    & 65          &4 & 6\\ \hline
5  & NATOPS         & 51       &24 & 6\\ \hline
6  & RacketSports       & 30        &6 & 4\\ 

\hline
\end{tabular}
\end{table}

\subsection{Implementation Details}
The classifier we used in this work is a fully convolutional neural network (FCN), originally proposed by Wang et al. \cite{wang2017time}. It is important to note that our M-CELS model is model-agnostic, meaning that any other deep learning model can be used as the classification model. We optimize M-CELS using the ADAM optimizer \cite{kingma2014adam} with a learning rate of 0.1 and train for 1000 epochs. Early stopping is implemented to conclude the optimization process when a satisfactory level of convergence on the loss function is achieved. The $\lambda$ we used in our experiments is set to 1. And the threshold $k$ is set to 0.5. The source code of this work is available on our GitHub repository \footnote{\url{https://github.com/Luckilyeee/M-CELS}}. This work also includes an ablation study and an evaluation of interpretability. Due to space constraints, these details have been made available on our project website. Please refer to \footnote{\url{https://sites.google.com/view/m-cels/home}} for more information.

\subsection{Evaluation metrics \label{Sec:ExpResults}}
The goal of our experiments is to assess the performance of the baseline methods concerning all the desired properties of an ideal counterfactual method. To evaluate our proposed method M-CELS, we compare our method with the other three baselines in terms of validity, proximity, and sparsity.

The first one is \textit{target probability}, which is used to evaluate the validity property. We define the \textbf{validity} metric by comparing the target class probability for the prediction of the counterfactual explanation result. The closer the target class probability is to 1, the better. 

The second evaluation metric we used is the \textit{L1 distance} to demonstrate the \textbf{proximity}, which is defined in Equation~\ref{equ:L1}. It measures the closeness between the generated counterfactual $\mathbf{x}'$ and the original instance of interest $\mathbf{x}$, a smaller L1 distance is preferred. 

\begin{equation}
   L1 \ distance = \sum_{d=1}^D\sum_{t=1}^T(|\mathbf{x}_{t,d}- \mathbf{x}_{t,d}'|), 
    \label{equ:L1} 
\end{equation}

Then we use the \textit{sparsity level} to evaluate the \textbf{sparsity}. Sparsity is evaluated by calculating the percentage of data points that remain unchanged after the perturbation. The highest sparsity is an indicator that the time series perturbations made in $\mathbf{x}$ to achieve $\mathbf{x}'$ are minimal. Therefore, a higher sparsity level is desirable. The equations designed by \cite{filali2022mining} are shown in \ref{equ:spar}-\ref{equ:func}. 
\begin{equation}\label{equ:spar}
    Sparsity = 1- \frac{\sum_{d=1}^D\sum_{t = 1}^Tg({\mathbf{x}'_{t,d}}, \mathbf{x}_{t,d})}{T} 
\end{equation}
\begin{equation}\label{equ:func}
    g(x, y)=\left\{\begin{array}{ll}
    1, & \text { if } x \neq y \\
    0, & \text { otherwise }
    \end{array}\right.
\end{equation}

% Finally, we compare the \textit{running time} of the counterfactual instances generation to verify the \textbf{efficiency}, the faster a valid counterfactual instance can be generated the better.

\subsection{Evaluation results}
\begin{figure*}[!htbp]
  \centering
  \begin{minipage}{0.33\linewidth}
    \centering
    \includegraphics[width=\linewidth]{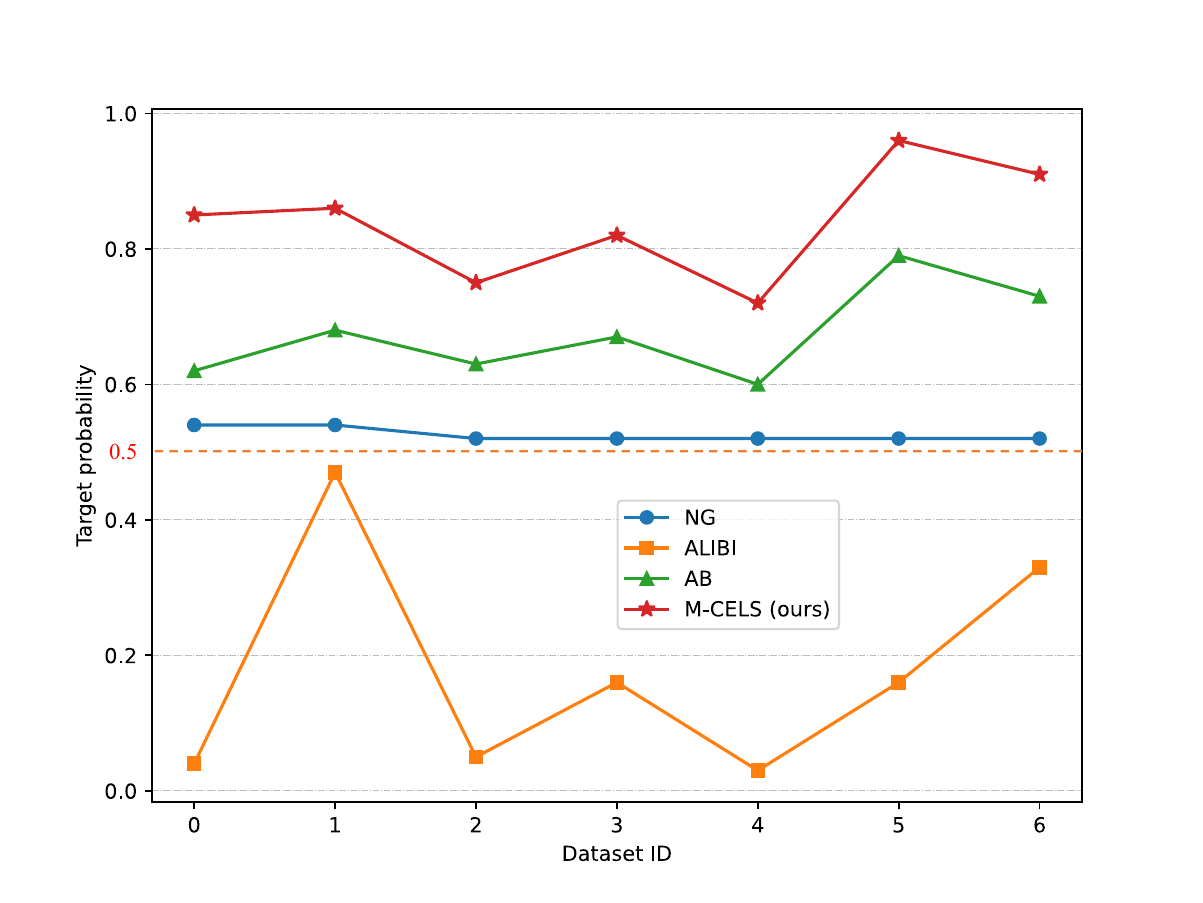}
    \subcaption{Target probability (the higher the better)}
    \label{fig:prob}
  \end{minipage}%
  \begin{minipage}{0.33\linewidth}
    \centering
    \includegraphics[width=\linewidth]{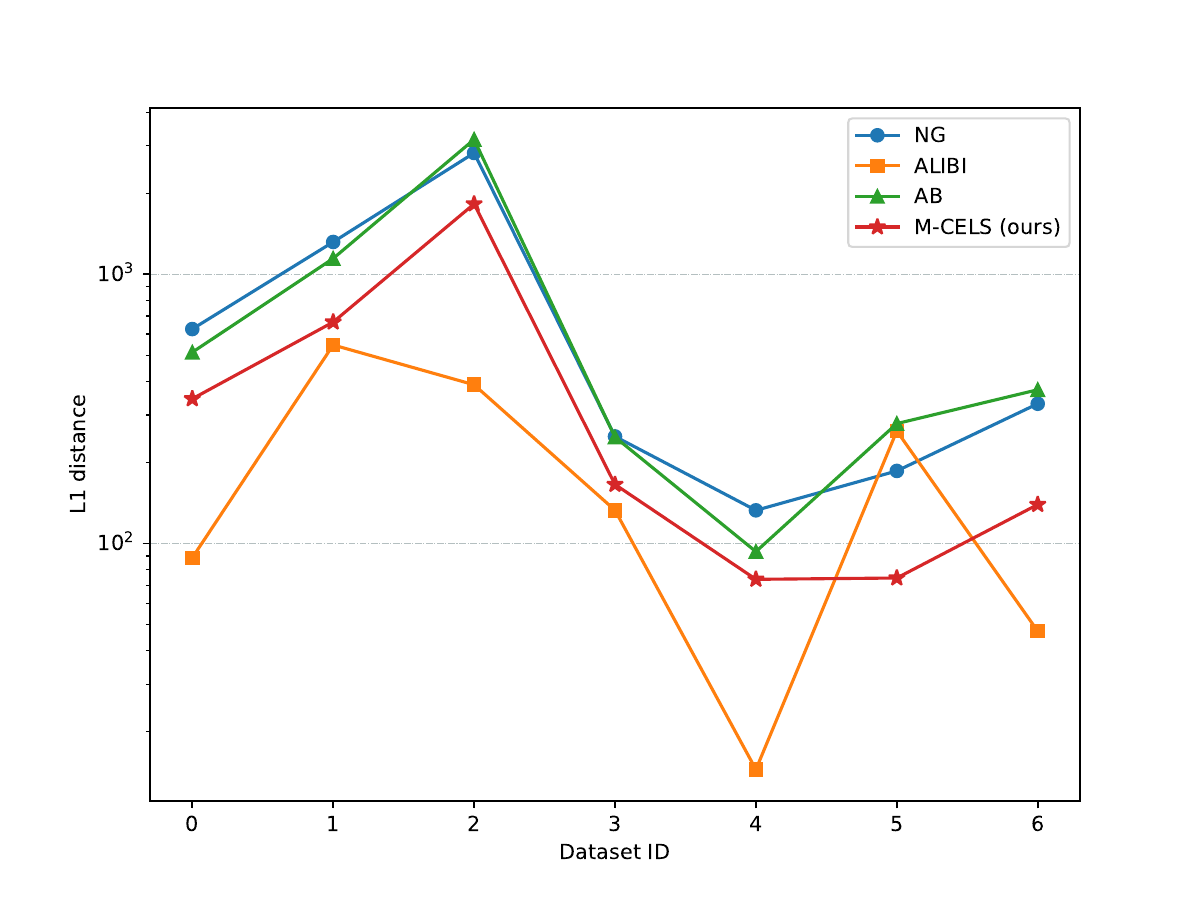}
    \subcaption{L1 distance (the lower the better)}
    \label{fig:L1}
  \end{minipage}
  \begin{minipage}{0.33\linewidth}
    \centering
    \includegraphics[width=\linewidth]{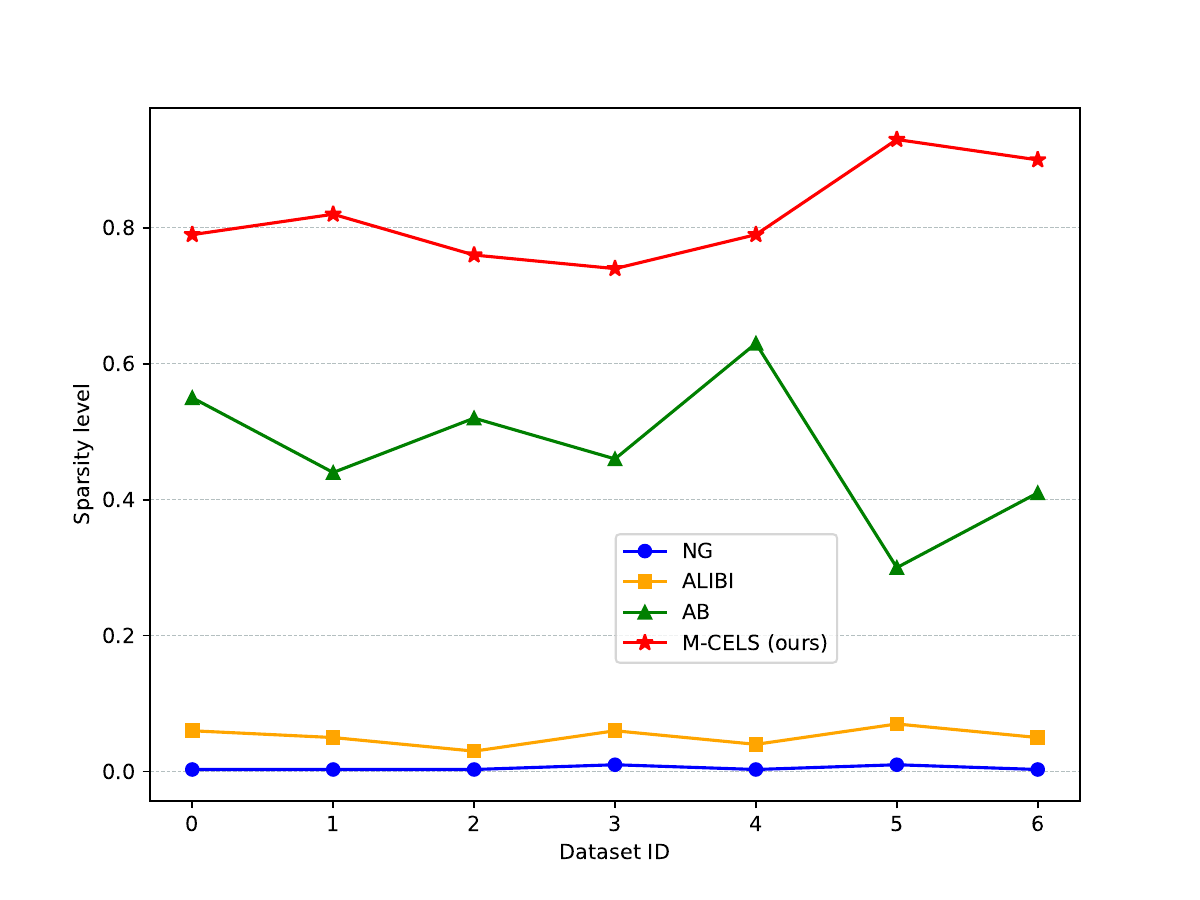}
    \subcaption{Sparsity level (the higher the better)}
    \label{fig:Sparsity}
  \end{minipage}%
  \caption{Evaluation results of CF explanations by ALIBI, NG-CF, AB-CF, and our proposed method. (All reported values are the average over the counterfactual set)}
  \label{fig:resultfig}
\end{figure*}

Figure \ref{fig:resultfig} presents the outcomes of our experiments, comparing our proposed method, M-CELS, against several baseline models. Each figure highlights a different aspect of performance, providing a comprehensive evaluation of the effectiveness of our approach.

In Figure \ref{fig:prob}, we depict the target probability of the generated counterfactuals. Notably, our approach achieves the highest target probability, indicating superior validity. In contrast, ALIBI's target probability falls below 50\%, suggesting invalid counterfactuals. NG achieves a target probability slightly over 50\%, which means the perturbation stops whenever the target probability exceeds 50\%. This indicates that while NG can generate valid counterfactuals, it does so just barely, often stopping at the minimal threshold. This suggests that our method provides more reliable and valid counterfactual explanations compared to the other baselines, ensuring that the generated counterfactuals align more closely with the desired outcomes.

Next, in Figure \ref{fig:L1}, we assess the proximity property using the L1 distance metric across all counterfactual explanation models. M-CELS achieves the second lowest L1 distance compared to NG, AB, and ALIBI. Although ALIBI exhibits the lowest L1 distance, its counterfactuals are considered invalid due to low target probabilities. This demonstrates that our method strikes a good balance between closeness to the original instance and the validity of the counterfactuals. The ability to generate counterfactuals that are both valid and proximate to the original data points is crucial for interpretability and practical applicability.

Moving on to Figure \ref{fig:Sparsity}, we compare the sparsity properties (the percentage of unchanged data points after perturbation) for generated counterfactuals across all models. M-CELS outperforms the other baselines in terms of sparsity, indicating minimal alterations to crucial data points during perturbation. High sparsity is essential as it preserves the integrity of the original data while ensuring that the counterfactual explanation remains meaningful and interpretable. By making fewer changes, M-CELS maintains the context and relevance of the original instance, making it easier for users to understand the necessary adjustments for a different outcome.

% Finally, Figure \ref{fig:time} compares the efficiency of counterfactual generation among all models. M-CELS requires more running time compared to NG and AB but less time compared to ALIBI. This suggests that while our method may take slightly longer to generate counterfactuals, it remains efficient compared to the most time-consuming baseline. The additional time required by M-CELS is justified by the superior validity, sparsity, and proximity of the counterfactuals it produces. This balance between efficiency and performance is critical for practical applications where both speed and quality of explanations are important. Although M-CELS requires more processing time compared to NG and AB, it remains more efficient than ALIBI while providing significantly more reliable and interpretable counterfactuals. 

In summary, counterfactual explanations generated by M-CELS exhibit superior performance in terms of validity, sparsity, and proximity. This comprehensive performance profile underscores the effectiveness of M-CELS in generating high-quality counterfactual explanations that are both meaningful and practical for real-world applications.

\section{Conclusion}
In this paper, we introduce M-CELS as an innovative solution that advances counterfactual explanations for multivariate time series classification. First, we present novel optimization techniques specifically designed to handle the complexities of multivariate time series, ensuring the generation of meaningful and interpretable counterfactuals across multiple dimensions. Second, M-CELS leverages learned saliency maps to target the most influential time steps, leading to focused perturbations and promoting higher sparsity. This results in fewer changes to the original instance, making the counterfactuals more actionable and easier to interpret. Finally, these targeted and sparse perturbations enhance the overall interpretability of time series classification, enabling users to better understand model decisions and the influence of key variables over time. Our experiments across seven widely used datasets show that M-CELS outperforms three state-of-the-art counterfactual models, delivering significantly sparser, more proximate, and more valid counterfactual explanations. As a future step, we aim to improve M-CELS' computational efficiency for better scalability on larger datasets and real-time applications.
% As a future step, we would like to explore the possibility of further improving the contiguity of the generated counterfactuals based on M-CELS
\section*{Acknowledgment}
This project has been supported in part by funding from CISE and GEO Directorates under NSF awards \#2204363, \#2240022, \#2301397, and \#2305781.

\bibliography{./mybib}
\bibliographystyle{IEEEtran}
\vspace{12pt}
\color{red}

\end{document}